\documentclass{article}

\usepackage[final]{neurips_2019}

\usepackage[utf8]{inputenc} 
\usepackage[T1]{fontenc}    
\usepackage{hyperref}       
\usepackage{url}            
\usepackage{booktabs}       
\usepackage{amsfonts}       
\usepackage{nicefrac}       
\usepackage{microtype}      

\usepackage{amsmath}
\usepackage{amssymb}        
\usepackage{bm}             

\usepackage{graphicx}       
\usepackage{subcaption}     
\usepackage{color,xcolor,colortbl}

\usepackage{tikz}
\usepackage{tikz-cd}
\usepackage{pgfplots}
\usepackage{flowchart}
\usetikzlibrary{arrows}
\pgfplotsset{compat=1.14}

\let\originalleft\left
\let\originalright\right
\renewcommand{\left}{\mathopen{}\mathclose\bgroup\originalleft}
\renewcommand{\right}{\aftergroup\egroup\originalright}

\title{BERTgrid: Contextualized Embedding for 2D Document Representation and Understanding}

\author{%
  Timo I. Denk\thanks{Equal contribution} \\
  SAP SE\\
  Machine Learning R\&D, Berlin Germany\\
  \texttt{mail@timodenk.com} \\
  \And
  Christian Reisswig\footnotemark[1] \\
  SAP SE \\
  Machine Learning R\&D, Berlin Germany\\
  \texttt{christian.reisswig@sap.com} \\
}

\begin{document}

\maketitle

\begin{abstract}
For understanding generic documents, information like font sizes, column layout, and generally the positioning of words may carry semantic information that is crucial for solving a downstream document intelligence task.
Our novel \textit{BERTgrid}, which is based on \textit{Chargrid} by \cite{DBLP:conf/emnlp/KattiRGBBHF18}, represents a document as a grid of contextualized word piece embedding vectors, thereby making its spatial structure and semantics accessible to the processing neural network.
The contextualized embedding vectors are retrieved from a BERT language model.
We use BERTgrid in combination with a fully convolutional network on a semantic instance segmentation task
for extracting fields from invoices. We demonstrate its performance on tabulated line item and document header field extraction.
\end{abstract}

\section{Introduction}
\label{sec:introduction}

Documents often come in a variety of layouts and formats. 
For instance, a single document may contain isolated text boxes, tabular arrangements, multiple columns, and different font sizes. This layout can carry crucial semantic information. 
In classical natural language processing (NLP), however, the layout information is completely discarded as the document text is simply a sequence of words.
Without access to the layout, a downstream task such as extraction of tabulated data can become much harder -- and in some cases impossible to solve -- since the necessary serialization may lead to severe information loss. Instead of working on the textual level, it is possible to directly apply methods from computer vision (CV)  (e.g.~\cite{DBLP:conf/nips/RenHGS15}) to work on the raw document pixel level which naturally retains the two-dimensional (2D) document structure. However, this is impractical, as a machine learning model would first need to learn textual information from the raw pixel data followed by the semantics.  

Recent approaches have designed a hybrid between NLP and CV methods for document intelligence: Chargrid (\cite{DBLP:conf/emnlp/KattiRGBBHF18}), followed more recently by CUTIE (\cite{cutie}), construct a 2D grid of characters or words from a document and feed it into a neural model, thereby preserving the spatial arrangement of the document. The symbols in the original document are embedded in some vector space, yielding a rank-3 tensor (width, height, embedding). Both papers report significant benefits of using such a grid approach over purely sequential 1D input representations, especially for semantically understanding tabulated or otherwise spatially arranged text like line items.

With our contribution \textit{BERTgrid}, we incorporate contextualized embedding into the grid document representation. More specifically, we use a BERT language model (\cite{DBLP:conf/naacl/DevlinCLT19}) pre-trained on a large pool of unlabeled documents from the target domain to compute contextualized feature vectors for every word piece in a document.
We demonstrate the effectiveness of BERTgrid on an invoice information extraction task from document tables and headers. We compare our results to Chargrid and find significant improvements from $61.76\%\pm0.72$ to $65.48\%\pm0.58$ on an invoice dataset previously described in \cite{DBLP:conf/emnlp/KattiRGBBHF18}.

\section{Method}
\label{sec:method}

\subsection{BERTgrid document representation}

Instead of constructing a grid on the character level and embedding each character with one-hot encoding as in \cite{DBLP:conf/emnlp/KattiRGBBHF18}, we construct a grid on the word-piece level and embed with dense contextualized vectors from a BERT language model. 

Formally, let a document be denoted by $\mathcal{D}:=\left\{\left(
        w^{(j)},x^{(j)}_\text{min},y^{(j)}_\text{min},x^{(j)}_\text{max},y^{(j)}_\text{max}
    \right)\mid
    j\in\left\{1,\dots,n\right\}\right\}\,,$
consisting of $n$ word pieces $w^{(j)}$, each of which is associated with a non-overlapping bounding box $x^{(j)}_\text{min},y^{(j)}_\text{min},x^{(j)}_\text{max},y^{(j)}_\text{max}$. Further, let $\mathcal{S}:=\begin{bmatrix}w^{(1)} & w^{(2)} & \dots & w^{\left(n\right)}\end{bmatrix}$ be the line-by-line serialized version of $\mathcal{D}$.
Using all word pieces $j\in\left\{1,\dots,n\right\}$, the BERTgrid representation of the document is defined as \begin{equation}\label{eq:bertgridtensor}
    \bm{W}_{x,y,:}:=\begin{cases}
        \bm{e}\left(\mathcal{S},j\right) & \text{if }x^{(j)}_\text{min}\le x\le x^{(j)}_\text{max}\land y^{(j)}_\text{min}\le y\le y^{(j)}_\text{max}\\
        \bm{0}_d & \text{otherwise}\,,
    \end{cases}
\end{equation} 
where $d$ is the embedding dimensionality, $\bm e$ is the embedding function, and $\bm{0}_d$ denotes an all-zero vector which we use for background. 
We implement $\bm{e}$ using a pre-trained BERT language model. During evaluation of $\bm{e}$, $\mathcal{S}$ is fed into the BERT model. The representation of the second-to-last hidden layer for the $j$th position is used to embed $w^{(j)}$ at position $x,y$ in $\bm{W}$. Fig.~\ref{fig:invoicegroundtruth}~(d) visualizes a BERTgrid tensor $\bm{W}$.

\subsection{Models}

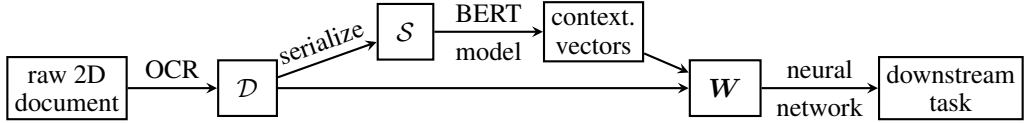
\begin{figure}
    \centering
    \begin{tikzpicture}[>=stealth, thick]
        \node (2ddocraw) at (0,0) [draw, process, align=flush center, inner sep=.1cm] {raw 2D\\document};
        \node (2ddoc) at (2.4,0) [draw, process, align=flush center, inner sep=.25cm] {$\mathcal{D}$};
        \node (1ddoc) at (4.5,.75) [draw, process, align=flush center, inner sep=.25cm] {$\mathcal{S}$};
        \node (contvecs) at (7,.75) [draw, process, align=flush center, inner sep=.1cm] {context.\\vectors};
        \node (bertgrid) at (8.75,0) [draw, process, align=flush center, inner sep=.25cm] {$\bm{W}$};
        \node (out) at (11.75,0) [draw, process, align=flush center, inner sep=.1cm] {downstream\\task};
        
        \draw [->] (2ddocraw) -- node [above] {OCR} (2ddoc);
        \draw [->] (2ddoc) -- node [above,sloped] {serialize} (1ddoc);
        \draw [->] (1ddoc) -- node [above,sloped] {BERT} node [below,sloped] {model} (contvecs);
        \draw [->] (contvecs) -- node [above,sloped] {} (bertgrid);
        \draw [->] (2ddoc) -- node [above,sloped] {} (bertgrid);
        \draw [->] (bertgrid) -- node [above,sloped] {neural} node [below,sloped] {network} (out);
    \end{tikzpicture}
    \caption{The BERTgrid pipeline. The OCR step is optional and not necessary for file formats which directly encode words and their positions (e.g.~HTML, DOC, or PDF).}
    \label{fig:bertgridcomponents}
\end{figure}

Our model pipeline is summarized in Fig.~\ref{fig:bertgridcomponents}. 
A raw document image is first passed through an OCR engine to retrieve the words and their positions, i.e.~$\mathcal{D}$.
We then serialize $\mathcal{D}$ resulting in $\mathcal{S}$ which is subsequently passed through a BERT language model ($\mathrm{BERT_{BASE}}$ configuration from \cite{DBLP:conf/naacl/DevlinCLT19}) to get a contextualized vector for each word. Together with positional information from $\mathcal{D}$, we construct $\mathbf{W}$ according to Eq.~\ref{eq:bertgridtensor}.
For our downstream information extraction task, we use the same fully convolutional encoder-decoder neural network architecture and the same semantic segmentation and bounding box regression training tasks as \cite{DBLP:conf/emnlp/KattiRGBBHF18}, except $\bm{W}$ is the input to the neural network.
Just like in \cite{DBLP:conf/emnlp/KattiRGBBHF18}, we obtain extracted document strings by comparing the predicted segmentation mask and bounding boxes with $\mathcal{D}$.
We interchangeably use BERTgrid for denoting just the document representation or the complete model consisting of input and network.

As an extension to BERTgrid, we also construct a second model {[C+BERTgrid]} which combines the Chargrid and BERTgrid input representations.
For that, we replicate the first convolutional block of the neural network to have a Chargrid and a BERTgrid branch. Both are subsequently merged by adding the two hidden representations.

We further create a {[Wordgrid]} model and likewise a {[C+Wordgrid]} model. The Wordgrid representation is identical to Chargrid in that it is non-contextualized. The embedding, however, is on the word level, learned with word2vec (\cite{DBLP:journals/corr/abs-1301-3781}) on the unlabeled invoices.
The {[Wordgrid]} representation replicates \cite{cutie} except it is initialized with semantically rich embedding from word2vec and not fine-tuned during training.

All models are trained for 800k iterations on a single Nvidia V100 GPU each. The BERT model with sequence length 512 is pre-trained for 2M steps and not fine-tuned on the downstream task.

\section{Experiments}
\label{sec:experiments}

\subsection{Data}

\begin{figure}
  \centering
  \begin{subfigure}[t]{0.24\textwidth}
    \centering
    \includegraphics[height=1.7in]{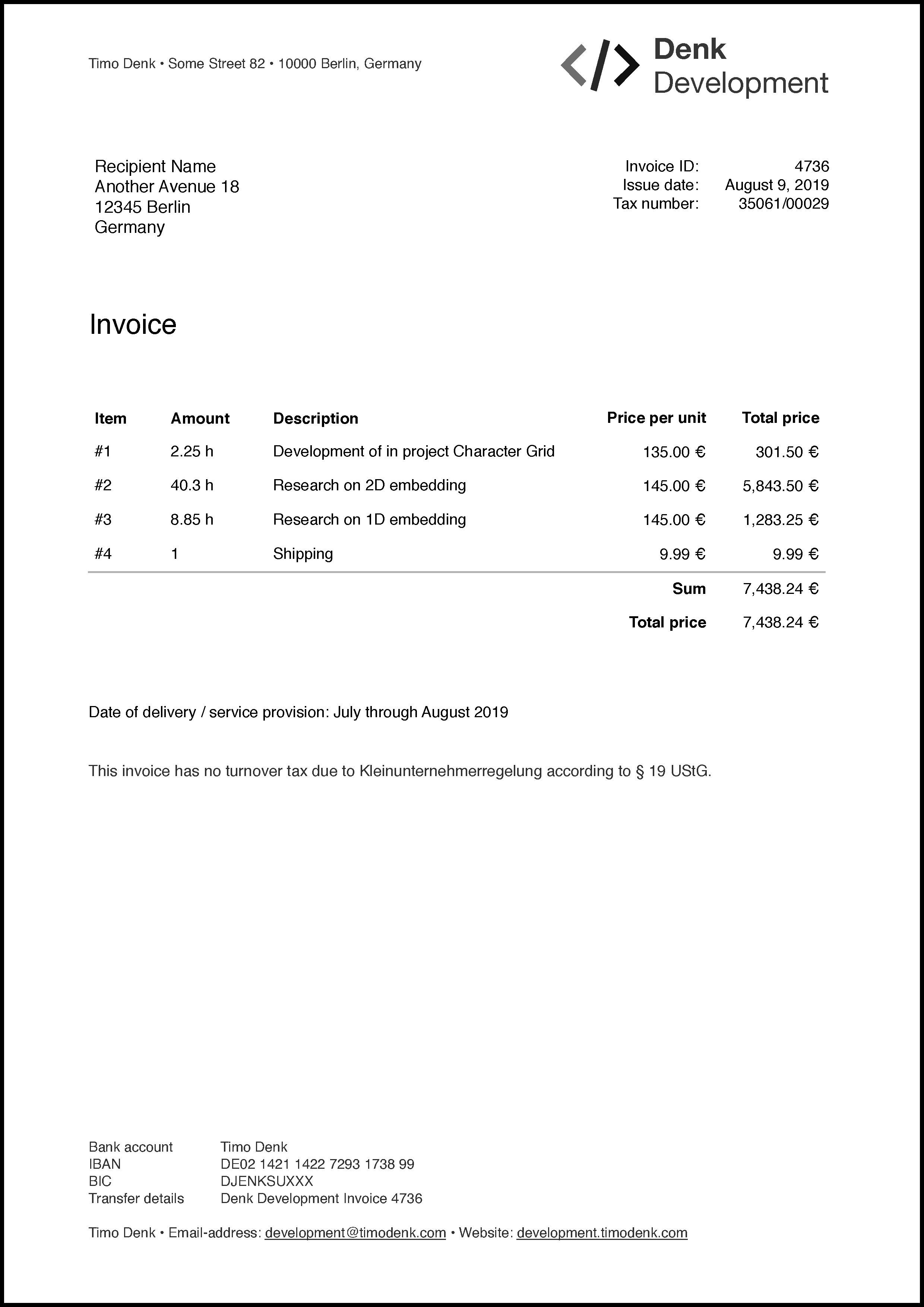}
    \caption{Invoice}
  \end{subfigure}%
  ~ 
  \begin{subfigure}[t]{0.24\textwidth}
    \centering
    \includegraphics[height=1.7in]{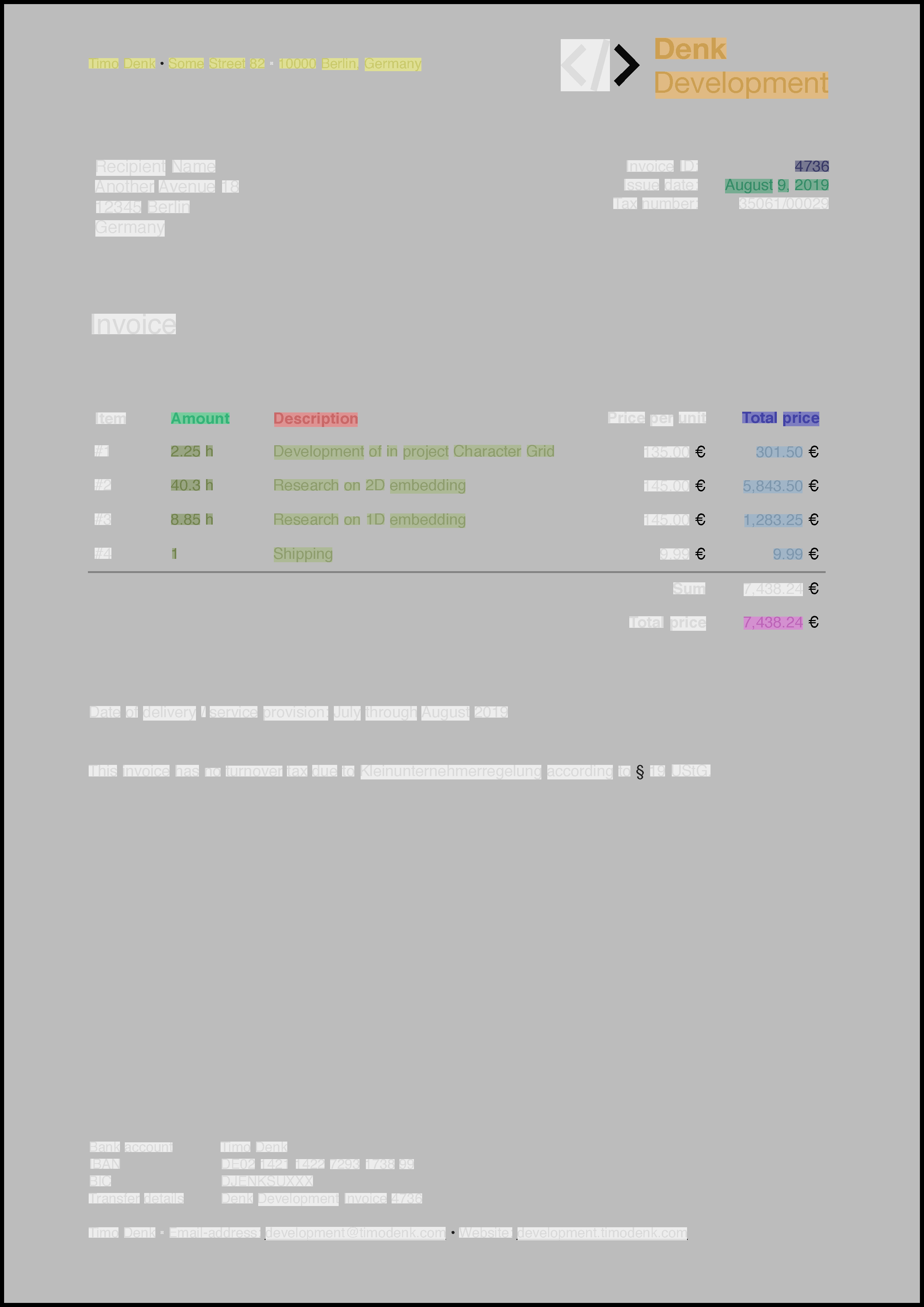}
    \caption{Segmentation mask}
  \end{subfigure}%
  ~ 
  \begin{subfigure}[t]{0.24\textwidth}
    \centering
    \includegraphics[height=1.7in]{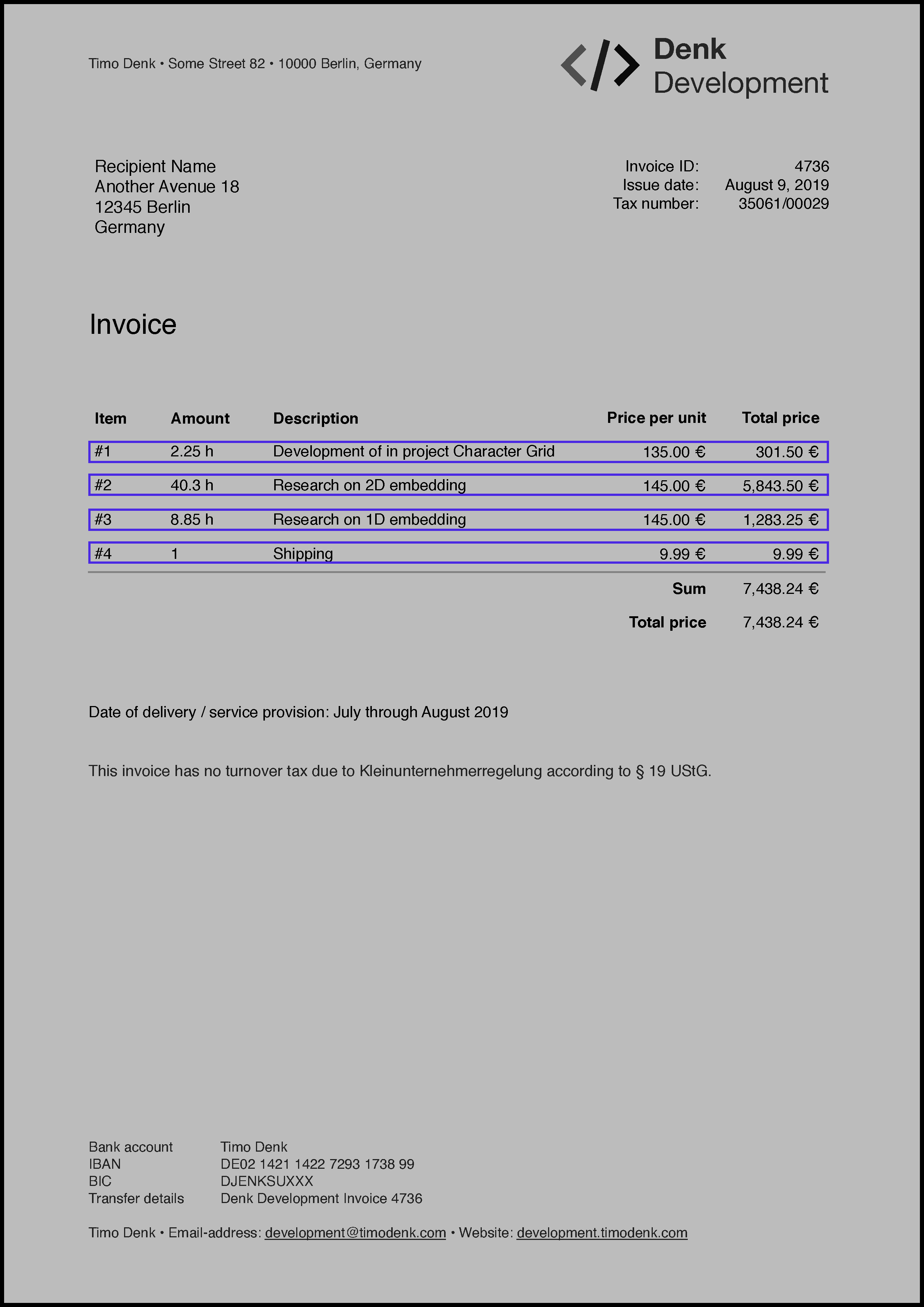}
    \caption{Bounding boxes}
  \end{subfigure}%
  ~ 
  \begin{subfigure}[t]{0.24\textwidth}
    \centering
    \includegraphics[height=1.7in]{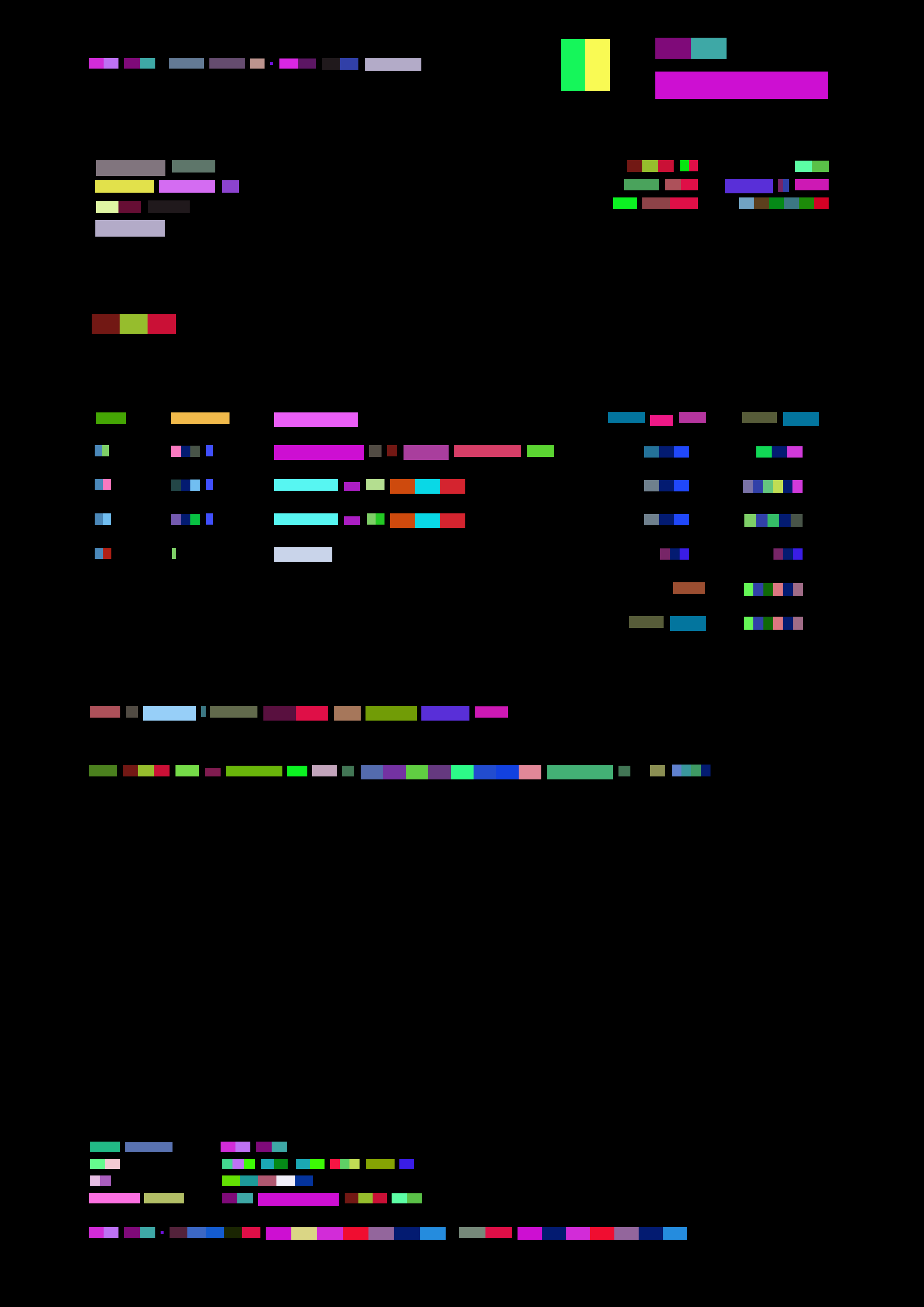}
    \caption{BERTgrid}
  \end{subfigure}
  \caption{Invoice dataset sample ground truth and BERTgrid. Each text field of (a) is annotated with a label (b). Line items are additionally annotated with bounding boxes (c). BERTgrid (d) embeds the text data on the word-piece level; different colors stand for different vectors.}
  \label{fig:invoicegroundtruth}
\end{figure}

As a concrete example for document intelligence, we extract key-value information from invoices without making assumptions on the invoice layout.
We distinguish two kinds of fields: (1) \textit{header fields} and (2) \textit{line item fields}. The former includes invoice amount, invoice number, invoice date, and vendor name and address. The latter includes line item quantity, description, VAT amount/rate, and total price. It is usually contained in tabulated form and can occur in multiple instances per invoice. For each line item, all associated fields are grouped by a single bounding box per line item. 
Not all fields are always present on an invoice.

We use the same dataset for training and testing as described in \cite{DBLP:conf/emnlp/KattiRGBBHF18}. It is comprised of 12k samples which we split 10k/1k/1k for training/validation/testing. The invoices are from a large variety of different vendors and the sets of vendors contained in training, validation, and testing samples are disjoint. Languages are mixed, with the majority being English.
An example invoice along with its ground truth annotations is shown in Fig.~\ref{fig:invoicegroundtruth}. 
In addition, we use a second, much larger dataset comprised of 700k unlabeled invoices.
This dataset is serialized resulting in about 800 MB of plain text data. We use it for pre-training BERT from scratch as well as learning embedding vectors with word2vec.

\subsection{Metric}

We use the evaluation metric from \cite{DBLP:conf/emnlp/KattiRGBBHF18}.
This measure is similar to the edit distance, a measure for the dissimilarity of two strings.
For a given field, we count the number of insertions, deletions, and modifications of the predicted instances to match the ground truth (pooled across the entire test set). The measure is computed as \begin{equation}\label{eq:editdist}
    1 - \frac{\#[\text{\footnotesize insertions}]+\#[\text{\footnotesize deletions}]+\#[\text{\footnotesize modifications}]}{N}\,,
\end{equation}
where $N$ is the total number of instances of a given field occurring in the ground truth of the entire test set.
This measure can be negative, meaning that it would be less work to perform the extraction manually.
The best value it can reach is $1$, corresponding to perfect extraction.

\subsection{Results and discussion}

\begin{table}
    \caption{Extraction accuracy on selected invoice header and line item (LI) fields for different document representation methods with the same neural network. The mean is computed across all fields. Contextualized representations consistently achieve the best extraction scores.}
    \label{tab:resultsheader}
    \centering
    {\footnotesize\begin{tabular}{lrrrrrrrr}
        \toprule
         & Mean & Amount & Number & Date & Vendor name & LI mean & LI quantity \\\midrule
{[Chargrid]} & 61.76\% & 91.42\% & 83.90\% & 85.74\% & 40.91\% & 56.59\% & 66.97\% \\
{[Wordgrid]} & 60.36\% & 88.79\% & 77.35\% & 84.08\% & 39.75\% & 55.98\% & 66.19\% \\
{[C+Wordgrid]} & 62.94\% & 90.53\% & 84.34\% & 87.12\% & 41.63\% & 58.19\% & 66.79\% \\
{[BERTgrid]} & 64.21\% & \textbf{92.44\%} & 84.99\% & 87.79\% & 44.86\% & 59.38\% & 71.97\% \\
{[C+BERTgrid]} & \textbf{65.48\%} & 92.38\% & \textbf{86.25\%} & \textbf{88.46\%} & \textbf{47.22\%} & \textbf{60.42\%} & \textbf{73.18\%} \\
        \bottomrule
    \end{tabular}}
\end{table}

Tab.~\ref{tab:resultsheader} shows the results in terms of the evaluation measure for different input representations. All results are averaged over four randomly initialized training runs.
\cite{DBLP:conf/emnlp/KattiRGBBHF18, cutie} have shown that grid-based approaches like {[Chargrid]} or {[Wordgrid]} outperform conventional sequential models as well as purely image-based methods, so we use {[Chargrid]} as our baseline, with $61.76\%\pm0.72$. 

We find {[Wordgrid]} consistently performs worse on all fields, which can be attributed to the high rate of out-of-vocabulary words.
Only when combining word- and character-level information in {[C+Wordgrid]}, we outperform the baseline on most fields.
Our contributions {[BERTgrid]} and {[C+BERTgrid]} significantly outperform all other models.
With a mean absolute performance of $65.48\%\pm0.58$, {[C+BERTgrid]} is $6.02\%$ (relative) above the baseline.

We assume the performance of BERTgrid stems from (i)~embedding on the word-piece level and (ii)~contextualization. 
Rather than learning to represent words first, the network directly gets access to semantically meaningful word(-piece)-level information. 
For instance, words such as \textit{avenue}, \textit{street}, and \textit{drive} are very different when embedded on the character level, but will be mapped to approximately the same embedding vector. 
We observe that both {[C+Wordgrid]} and {[C+BERTgrid]} converge faster than {[Chargrid]} which supports this statement.

During language model pre-training on the large, unlabeled dataset, knowledge about the language of invoices is distilled into the BERT model parameters. Compared to simpler, non-contextualized embedding methods such as word2vec, it has sufficient capacity to capture complex dependencies. This distilled knowledge is made accessible via the BERTgrid representation and eases the downstream task significantly.

We acknowledge the BERT model has only access to $\mathcal{S}$, not $\mathcal{D}$. Future work could use 2D positional encodings to preserve the layout structure also during language model pre-training and inference.

\section{Conclusions}
\label{sec:conclusion}

We introduced BERTgrid, a generic, contextualized, grid-based representation method for 2D documents. 
It embeds word pieces with contextualized vectors in a 2D grid and it utilizes unlabeled data which is often available in abundance.
This representation eases the learning of downstream tasks such as key-value extraction from invoices where the layout carries semantically important information. We have shown that BERTgrid is superior to other approaches, especially to a Chargrid and a Wordgrid baseline, which have previously been shown to outperform sequential models, see \cite{DBLP:conf/emnlp/KattiRGBBHF18, cutie}. 

{\small
\bibliographystyle{rusnat}
\bibliography{bibliography}

\begin{thebibliography}{5}
\providecommand{\natexlab}[1]{#1}
\providecommand{\EM}{\em}
\providecommand{\RNtxt}{\relax}
\RNtxt{}

\bibitem[Devlin et~al.(2019)J.~Devlin, M.~Chang, K.~Lee,
  K.~Toutanova]{DBLP:conf/naacl/DevlinCLT19}
{\EM Devlin Jacob, Chang Ming{-}Wei, Lee Kenton, Toutanova Kristina}.
\newblock {BERT:} Pre-training of Deep Bidirectional Transformers for Language
  Understanding \allowbreak\newblock// 2019 Conference of the North American
  Chapter of the Association for Computational Linguistics: Human Language
  Technologies. 2019.

\bibitem[Katti et~al.(2018)A.~R. Katti, C.~Reisswig, C.~Guder, S.~Brarda,
  S.~Bickel, J.~H{\"{o}}hne, J.~B. Faddoul]{DBLP:conf/emnlp/KattiRGBBHF18}
{\EM Katti Anoop~R., Reisswig Christian, Guder Cordula, Brarda Sebastian,
  Bickel Steffen, H{\"{o}}hne Johannes, Faddoul Jean~Baptiste}.
\newblock Chargrid: Towards Understanding 2D Documents \allowbreak\newblock//
  2018 Conference on Empirical Methods in Natural Language Processing. 2018.

\bibitem[Mikolov et~al.(2013)T.~Mikolov, K.~Chen, G.~Corrado,
  J.~Dean]{DBLP:journals/corr/abs-1301-3781}
{\EM Mikolov Tomas, Chen Kai, Corrado Greg, Dean Jeffrey}.
\newblock Efficient Estimation of Word Representations in Vector Space
  \allowbreak\newblock// 1st International Conference on Learning
  Representations, {ICLR}. 2013.

\bibitem[Ren et~al.(2015)S.~Ren, K.~He, R.~B. Girshick,
  J.~Sun]{DBLP:conf/nips/RenHGS15}
{\EM Ren Shaoqing, He~Kaiming, Girshick Ross~B., Sun Jian}.
\newblock Faster {R-CNN:} Towards Real-Time Object Detection with Region
  Proposal Networks \allowbreak\newblock// Annual Conference on Neural
  Information Processing Systems 2015. 2015.

\bibitem[Zhao et~al.(2019)X.~Zhao, Z.~Wu, X.~Wang]{cutie}
{\EM Zhao Xiaohui, Wu~Zhuo, Wang Xiaoguang}.
\newblock {CUTIE:} Learning to Understand Documents with Convolutional
  Universal Text Information Extractor \allowbreak\newblock// CoRR. 2019.
  abs/1903.12363.

\end{thebibliography}
}

\end{document}